\documentclass{article}
\usepackage{spconf,amsmath,graphicx}
\usepackage{booktabs,subcaption,amsfonts,dcolumn}
\usepackage{multirow}
\usepackage{xcolor}

\title{Improving Spoken Language Understanding By Exploiting ASR N-best Hypotheses}
\name{Mingda Li$^{\star}$ $^{\ddagger}$ \thanks{$^{\ddagger}$ This work was done while the first author was an intern at Amazon.}, Weitong Ruan$^{\dagger}$, Xinyue Liu$^{\dagger}$, Luca Soldaini$^{\dagger}$, Wael Hamza$^{\dagger}$, Chengwei Su$^{\dagger}$ }
\address{$^{\star}$ University of California, Los Angeles, USA \\ $^{\dagger}$ Amazon Alexa AI, USA }
\begin{document}
%
\maketitle
\begin{abstract}
In a modern spoken language understanding (SLU) system, the natural language understanding (NLU) module takes interpretations of a speech from the automatic speech recognition (ASR) module as the input. The NLU module usually uses the first best interpretation of a given speech in downstream tasks such as domain and intent classification. However, the ASR module might misrecognize some speeches and the first best interpretation could be erroneous and noisy. Solely relying on the first best interpretation could make the performance of downstream tasks non-optimal. To address this issue, we introduce a series of simple yet efficient models for improving the understanding of semantics of the input speeches by collectively exploiting the $n$-best speech interpretations from the ASR module. 
\end{abstract}
\section{Introduction}
\label{sec:intro}
Currently, voice-controlled smart devices are widely used in multiple areas to fulfill various tasks, e.g. playing music, acquiring weather information and booking tickets. 
The SLU system employs several modules to enable the understanding of the semantics of the input speeches. 
When there is an incoming speech, the ASR module picks it up and attempts to transcribe the speech.
An ASR model could generate multiple interpretations for most speeches, which can be ranked by their associated confidence scores.
Among the $n$-best hypotheses, the top-1 hypothesis is usually transformed to the NLU module for downstream tasks such as domain classification, intent classification and named entity recognition (slot tagging). Multi-domain NLU modules are usually designed hierarchically \cite{tur2011spoken}. For one incoming utterance, NLU modules will firstly classify the utterance as one of many possible domains and the further analysis on intent classification and slot tagging will be domain-specific.

In spite of impressive development on the current SLU pipeline, the interpretation of speech could still contain errors.
Sometimes the top-1 recognition hypothesis of ASR module is ungrammatical or implausible and far from the ground-truth transcription \cite{peng2013search, jyothi2012large}. 
Among those cases, we find one interpretation exact matching with or more similar to transcription can be included in the remaining hypotheses ($2^{nd}- n^{th}$). 

To illustrate the value of the $2^{nd}- n^{th}$ hypotheses, we count the frequency of exact matching and more similar (smaller edit distance compared to the $1^{st}$ hypothesis) to  transcription for different positions of the $n$-best hypotheses list.
Table \ref{tbl:nbest} exhibits the results. For the explored dataset, we only collect the top 5 interpretations for each utterance ($n = 5$). Notably, when the correct recognition exists among the 5 best hypotheses, 50\% of the time (sum of the first row's percentages) it occurs among the $2^{nd}-5^{th}$ positions. Moreover, as shown by the second row in Table \ref{tbl:nbest}, compared to the top recognition hypothesis, the other hypotheses can sometimes be more similar to the transcription.

\begin{table}[h]
\vspace{-1.5ex}
    \captionof{table}{Spoken recognition quality distribution of the $n$ best hypotheses.}
    \label{tbl:nbest}
    \centering
    \begin{tabular}{|c|c c c c|} 
    \hline
        $n$ Best Rank Position & $2^{nd}$ & $3^{rd}$ & $4^{th}$ & $5^{th}$ \\
    \hline
        Match&19\%&14\%&10\%&7\%\\
        Prob (better than $1^{st}$ best) & 22\%&17\%&16\%&15\%\\
    \hline
    \end{tabular}
\vspace{-2ex}
\end{table}

Over the past few years, we have observed the success of reranking the $n$-best hypotheses 
\cite{peng2013search, charniak2005coarse, morbini2012reranking,  dikici2012classification, Sak2011DiscriminativeRO, sak2010fly,discriminative, collins2005discriminative, chan2004improving}
before feeding the best interpretation to the NLU module. These approaches propose the reranking framework by involving morphological, lexical or syntactic features \cite{discriminative, collins2005discriminative, chan2004improving}, speech recognition features like confidence score \cite{peng2013search, morbini2012reranking}, and other features like number of tokens, rank position \cite{peng2013search}.  They are effective to select the best from the hypotheses list and reduce the word error rate (WER) \cite{oba2007approach} of speech recognition. 

Those reranking models could benefit the first two cases in Table \ref{tbl:expMotiv} when there is an utterance matching with transcription. However, in other cases like the third row, it is hard to integrate the fragmented information  in multiple hypotheses. 

This paper proposes various methods integrating $n$-best hypotheses to tackle the problem. To the best of our knowledge, this is the first study that attempts to collectively exploit the $n$-best speech interpretations in the SLU system.  This paper serves as the basis of our $n$-best-hypotheses-based SLU system, focusing on the methods of integration for the hypotheses. 
Since further improvements of the integration framework require considerable setup and descriptions, where jointly optimized tasks (e.g. transcription reconstruction) trained with multiple ways (multitask \cite{caruana1997multitask}, multistage learning \cite{gong2013multi}) and more features (confidence score, rank position, etc.) are involved, we leave those to a subsequent article.





\begin{table}[t!]
    \captionof{table}{Motivating example: comparison of ASR $n$-Best hypotheses with the corresponding transcription. }
    \label{tbl:expMotiv}
    \vspace{-1ex}
    \scalebox{0.85}{
    \centering
    \begin{tabular}{|c|c|c|c|}
    \hline
    
    Transcription& $1^{st}$ best &$2^{nd}$ best&$3^{rd}$ best \\
    \hline
        \textbf{play muse}&play news&\textbf{play muse}&play mus\\
        \textbf{track on bose}&check on bowls& check on bose &\textbf{track on bose}\\
        \textbf{harry porter} & how \textbf{porter}& how patter&\textbf{harry} power\\
    \hline
            
    \end{tabular}

    }
\vspace{-3ex}
\end{table}

This paper is organized as follows. Section \ref{sec:pretrain} introduces the Baseline, Oracle and Direct models. Section \ref{sec:models} describes proposed ways to integrate $n$-best hypotheses during training. The experimental setup and results are described in Section \ref{sec:exp}. Section \ref{sec:conclusion} contains conclusions and future work.

\section{Baseline, Oracle and Direct Models}
\label{sec:pretrain}
 
\subsection{Baseline and Oracle}

\begin{figure}[!htp]
\vspace{-1ex}
	   \includegraphics[width=0.48\textwidth]{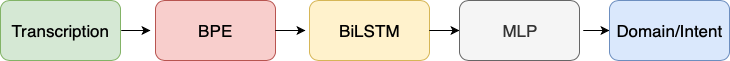}
        \caption{Baseline pipeline for domain or intent classification.}
        \label{fig:traditional}
        \vspace{-2ex}
\end{figure}

The preliminary architecture is shown in Fig. \ref{fig:traditional}. For a given transcribed utterance, it is firstly encoded with Byte Pair Encoding (BPE) \cite{sennrich2015neural}, a compression algorithm splitting words to fundamental subword units (\textit{pairs of bytes} or \textit{BP}s) and reducing the embedded vocabulary size. Then we use a BiLSTM \cite{schuster1997bidirectional} encoder and the output state of the BiLSTM is regarded as a vector representation for this utterance. Finally, a fully connected Feed-forward Neural Network (FNN) followed by a softmax layer, labeled as a multilayer perceptron (MLP) module, is used to perform the domain/intent classification task based on the vector.


For convenience, we simplify the whole process in Fig.\ref{fig:traditional} as a mapping $BM$ (Baseline Mapping) from the input utterance $S$ to an estimated tag's probability $p(\Tilde{t})$, where $p(\Tilde{t}) \leftarrow BM(S)$.
The $Baseline$ is trained on transcription and evaluated on ASR $1^{st}$ best hypothesis ($S=\text{ASR}\ 1^{st}\  \text{best})$. The $Oracle$ is trained on transcription and evaluated on transcription ($S = \text{Transcription}$). We name it Oracle simply because we assume that hypotheses are noisy versions of transcription. 
\subsection{Direct Models}
\label{subsec:combination}

Besides the Baseline and Oracle, where only ASR 1-best\footnote{We use ASR \textit{$n$-best hypotheses} or $n$-bests to denote the top $n$ interpretations of a speech, and the \textit{1,5-best} standing for the top 1 or 5 hypotheses.} 
hypothesis is considered, we also perform experiments to utilize ASR $n$-best hypotheses during evaluation. The models evaluating with $n$-bests and a BM (pre-trained on transcription) are called \textit{Direct Models} (in Fig. \ref{fig:eval}):

\begin{figure}[!htp]
	   \includegraphics[width=0.48\textwidth]{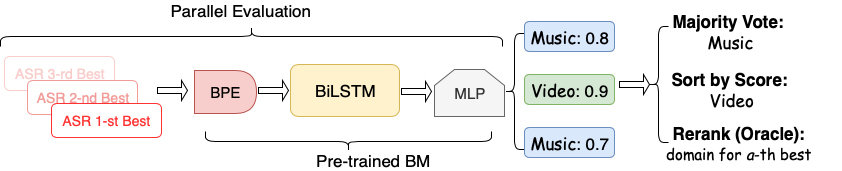}
       \vspace{-4ex}
        \caption{Direct models evaluation pipeline.}
        \vspace{-4ex}
        \label{fig:eval}
\end{figure}

\begin{itemize}
	\item \textit{Majority Vote.} We apply the BM model on each hypothesis independently and combine the predictions by picking the majority predicted label, i.e. Music.
	\vspace{-1ex}
	\item \textit{Sort by Score.} After parallel evaluation on all hypotheses, sort the prediction by the corresponding confidence score and choose the one with the highest score, i.e. Video.
	\vspace{-1ex}
	\item \textit{Rerank (Oracle).} 
	Since the current rerank models (e.g., \cite{peng2013search, charniak2005coarse, morbini2012reranking}) attempt to select the hypothesis most similar to transcription, we propose the Rerank (Oracle), which picks the hypothesis with the smallest edit distance to transcription (assume it is the $a$-th best) during evaluation and uses its corresponding prediction. 

\end{itemize}


\section{Integration of N-BEST Hypotheses}
\label{sec:models}
\begin{figure}[t!]
	   \includegraphics[width=0.48\textwidth]{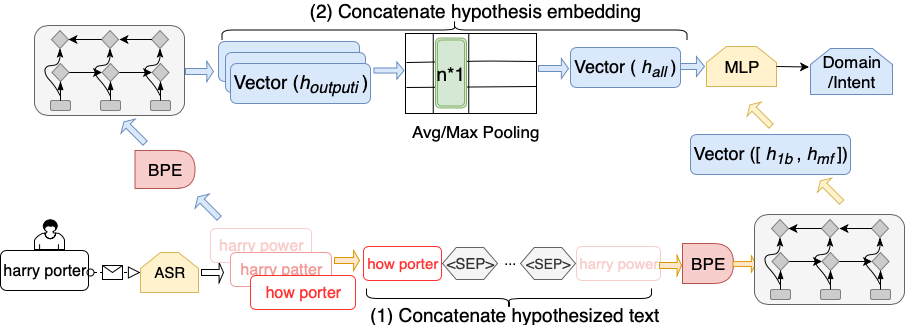}
       \vspace{-4ex}
        \caption{Integration of $n$-best hypotheses with two possible ways: 1) concatenate hypothesized text  and 2) concatenate hypothesis embedding.}
        \vspace{-2.9ex}
        \label{fig:integration}
\end{figure}

All the above mentioned models apply the BM trained on one interpretation (transcription). Their abilities to take advantage of multiple interpretations are actually not trained. As a further step, we propose multiple ways to integrate the $n$-best hypotheses during training. The explored methods can be divided into two groups as shown in Fig. \ref{fig:integration}. Let $H_1, H_2,..., H_n $ denote all the hypotheses from ASR and $bp_{H_k, i} \in BPs$ denotes the $i$-th pair of bytes (BP) in the $k^{th}$ best hypothesis. The model parameters associated with the two possible ways both contain: embedding $e_{bp}$ for pairs of bytes, BiLSTM parameters $\theta$ and MLP parameters $W, b$.

\subsection{Hypothesized Text Concatenation}
The basic integration method (\textit{Combined Sentence}) concatenates the $n$-best hypothesized text. We separate hypotheses with a special delimiter ($<$SEP$>$). We assume BPE totally produces $m$ BPs (delimiters are not split during encoding). Suppose the $n^{th}$ hypothesis has $j$ pairs. The entire model can be formulated as:
\begin{equation}
\small{
(h_1, ... , h_m)\leftarrow BiLSTM_{\theta}(bp_{H_1, 1},...,bp_{<sep>},...,bp_{H_n, j})
}
\label{bilstm}
\end{equation}
\begin{equation}
 \small{
 p(\Tilde{t}) = \sigma(W[h_{1b}, h_{mf}] +b)
 }
 \label{mlp}
 \end{equation}
 In Eqn. \ref{bilstm}, the connected hypotheses and separators are encoded via BiLSTM to a sequence of hidden state vectors. Each hidden state vector, e.g. $h_1$, is the concatenation of forward $h_{1f}$ and backward $h_{1b}$ states. The concatenation of the last state of the forward and backward LSTM forms the output vector of BiLSTM (concatenation denoted as $[,]$). Then, in Eqn. \ref{mlp}, the MLP module defines the probability of a specific tag (domain or intent)  $\Tilde{t}$ as the normalized activation ($\sigma$) output  after linear transformation of the output vector.
 

\subsection{Hypothesis Embedding Concatenation}
The concatenation of hypothesized text leverages the $n$-best list by transferring information among hypotheses in an embedding framework, BiLSTM. However, since all the layers have access to both the preceding and subsequent information, the embedding among $n$-bests will influence each other, which confuses the embedding and makes the whole framework sensitive to the noise in hypotheses.

As the second group of integration approaches, we develop models, \textit{PoolingAvg/Max}, on the concatenation of hypothesis embedding, which isolate the embedding process among hypotheses and summarize the features by a pooling layer. For each hypothesis (e.g., $i^{th}$ best in Eqn. \ref{bilstm_1} with $j$ pairs of bytes), we could get a sequence of hidden states from BiLSTM and obtain its final output state by concatenating the first and last hidden state ($h_{output_i}$ in Eqn. \ref{output}).
Then, we stack  all the output states vertically as shown in Eqn. \ref{outputs}. Note that in the real data, we will not always have a fixed size of hypotheses list. For a list with $r$ ($<n$) interpretations, we get the embedding for each of them and pad with the embedding of the first best hypothesis until a fixed size $n$. When $r\geq n$, we only stack the top $n$ embeddings. We employ $h_{output_1}$ for padding to enhance the influence of the top 1 hypothesis, which is more reliable. Finally, one unified representation could be achieved via    \textit{Pooling} (Max/Avg pooling with $n$ by 1 sliding window and stride 1)  on the concatenation and one score could be produced per possible tag for the given task. 
\begin{equation}
\small{
(h_{H_i, 1}, ... , h_{H_i, j})\leftarrow BiLSTM_{\theta}(bp_{H_i, 1},...,bp_{H_i, j})
}
\label{bilstm_1}
\end{equation}
\begin{equation}
\small{
h_{output_i} = [h_{H_i, 1b},\ h_{H_i, jf}]
\label{output}
}
\end{equation}
\begin{equation}
\small{
h_{outputs} = \begin{Bmatrix}

\begin{Bmatrix}
h_{output_1}\\...\\ h_{output_r}
\end{Bmatrix} & r-bests
\\
\begin{Bmatrix}
h_{output_1}\\ ...
\end{Bmatrix} & Padding\ with\ h_{output_1}
\end{Bmatrix}
}
\label{outputs}
\end{equation}
\begin{equation}
\small{
h_{all} = Pooling(h_{outputs})
}
\label{pooling}
\end{equation}
\begin{equation}
\small{
 p(\Tilde{t}) = \sigma(Wh_{all} +b)
 }
 \label{mlp_1}
 \end{equation}
\section{Experiment}
\label{sec:exp}
\subsection{Dataset}
We conduct our experiments on $\sim$ 8.7M annotated anonymised user utterances. They are annotated and derived from requests across 23 domains.
\subsection{Performance on Entire Test Set}

Table \ref{tbl:entire dataset} shows the relative error reduction (RErr)\footnote{The RErr for a model $m$ is calculated by comparing the relative difference between $100\% - MicroF1_{m}$ and $100\% - MicroF1_{\text{Baseline}}$.  } of Baseline, Oracle and our proposed models on the entire test set ($\sim$ 300K utterances) for multi-class domain classification. 
We can see among all the direct methods, predicting based on the hypothesis most similar to the transcription (Rerank (Oracle)) is the best.



\begin{table}[h]
\centering
\captionof{table}{Micro and Macro F1 score for multi-class domain classification.}
\vspace{-1ex}
\label{tbl:entire dataset}
\scalebox{0.8}{
 \begin{tabular}{ c|c|c} 
 \hline
Category & Model & RErr(\%)  \\
\hline
\multicolumn{2}{c|}{Baseline}& 	\textbf{0.00} \\
\hline
\multirow{3}*{Integration}&PoolingAvg	& \textbf{14.29} \\
&PoolingMax		& 13.20 \\
&Combined Sentence		& 11.67 \\
\hline
\multirow{3}*{Direct}&Sort by Score		&1.85\\
&Majority Vote	&	1.64\\
&Rerank (Oracle)	&	\textbf{3.71}   \\
\hline
\multicolumn{2}{c|}{Oracle}		&27.04\\
\hline
\end{tabular}}
\vspace{-2ex}
\end{table}

As for the other models attempting to integrate the $n$-bests during training, PoolingAvg gets the highest relative improvement, 14.29\%. It as well turns out that all the integration methods outperform direct models drastically. This shows that having access to $n$-best hypotheses during training is crucial for the quality of the predicted semantics. 
\subsection{Performance Comparison among Various Subsets}

\begin{table}[ht]
\centering
\captionof{table}{Performance comparison for the subset ($\sim$ 19\%) where ASR first best disagrees with transcription.}
\vspace{-1ex}
\label{tbl:disagree}
\scalebox{0.8}{
 \begin{tabular}{ c|c|c} 
 \hline
Category & Model &  RErr(\%)  \\
\hline
\multicolumn{2}{c|}{Baseline}&	\textbf{0.00}\\
\hline
\multirow{3}*{Integration}&PoolingAvg&	24.67\\
&PoolingMax	&	26.23\\
&Combined Sentence&	19.23\\
\hline
\multirow{3}*{Direct}&Sort by Score	&	9.95\\
&Majority Vote&	7.59\\
&Rerank (Oracle)&	7.25\\
\hline
\multicolumn{2}{c|}{Oracle}	&53.02\\

\hline
\end{tabular}}
\end{table}

\begin{table}[h]
\centering
\captionof{table}{Performance comparison for the subset ($\sim$ 81\%) where ASR first best agrees with transcription.}
\vspace{-1ex}
\label{tbl:agree}
\scalebox{0.8}{
 \begin{tabular}{ c|c |c} 
 \hline
Category & Model & RErr(\%) \\
\hline
\multicolumn{2}{c|}{Baseline}	&	\textbf{0.00}\\
\hline
\multirow{3}*{Integration}&PoolingAvg &	3.56\\
&PoolingMax										&-0.38\\
&Combined Sentence									&	4.50\\
\hline
\multirow{3}*{Direct}&Sort by Score	&	-8.269\\
&Majority Vote		&	-3.19\\
&Rerank (Oracle)	&	0.00\\

\hline
\multicolumn{2}{c|}{Oracle}	&	0.00\\
\hline
\end{tabular}}
\vspace{-3ex}
\end{table}

To further detect the reason for improvements, we split the test set into two parts based on whether ASR first best agrees with transcription and evaluate separately. Comparing Table \ref{tbl:disagree} and Table \ref{tbl:agree}, obviously the benefits of using multiple hypotheses are mainly gained when  ASR $1^{st}$ best disagrees with the transcription. When ASR $1^{st}$ best agrees with transcription, the proposed integration models can also keep the performance. Under that condition, we can still improve a little (3.56\%) because, by introducing multiple ASR hypotheses, we could have more information and when the transcription/ASR $1^{st}$ best does not appear in the training set's transcriptions, its $n$-bests list may have similar hypotheses included in the training set's $n$-bests. Then, our integration model trained on $n$-best hypotheses as well has clue to predict. 
The series of comparisons reveal that our approaches integrating the hypotheses are robust to the ASR errors and whenever the ASR model makes mistakes, we can outperform more significantly. 

\subsection{Improvements on Different Domains and Different Numbers of Hypotheses}
\begin{figure}[h]
\centering
\scalebox{0.9}{
       \includegraphics[width=0.45\textwidth]{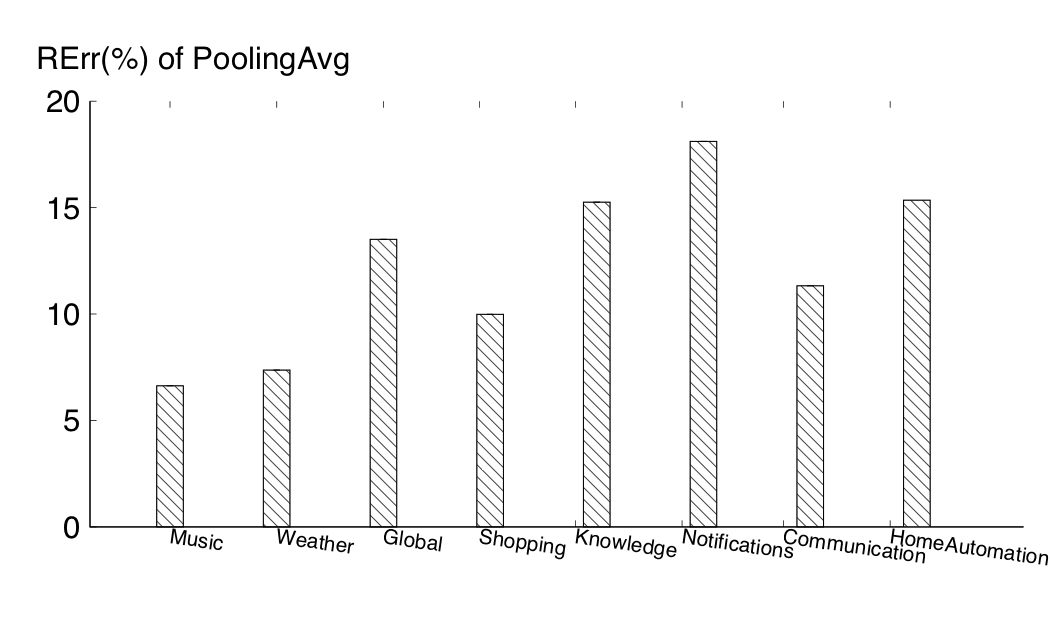}
       }
       \vspace{-3ex}
        \caption{Improvements on important domains.}
        \label{fig:important}
        \vspace{-1ex}

\end{figure}

Among all the 23 domains, we choose 8 popular domains for further comparisons between the Baseline and the best model of Table \ref{tbl:entire dataset}, PoolingAvg. Fig. \ref{fig:important} exhibits the results. We could find the PoolingAvg consistently improves the accuracy for all 8 domains.

\begin{figure}[h]
\centering
\vspace{-2ex}
\scalebox{0.5}{
       \includegraphics[width=0.5\textwidth]{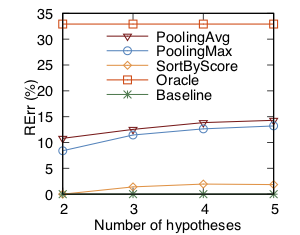}
       }
       \vspace{-1ex}
        \caption{The influence of different amount of hypotheses.}
         \vspace{-1.5ex}
        \label{fig:amount}

\end{figure}

In the previous experiments, the number of utilized hypotheses for each utterance during evaluation is five, which means we use the top 5 interpretations when the size of ASR recognition list is not smaller than 5 and use all the interpretations otherwise. Changing the number of hypotheses while evaluation, Fig. \ref{fig:amount} shows a monotonic increase with the access to more hypotheses for the PoolingAvg and PoolingMax (Sort by Score is shown because it is the best achievable direct model while the Rerank (Oracle) is not realistic). The growth becomes gentle after four hypotheses are leveraged. 

\subsection{Intent Classification}
\begin{table}[!htp]
\captionof{table}{Intent classification for three important domains.}
\vspace{-1ex}
    \label{tbl:intent}
    \centering
        \scalebox{0.8}{
         \begin{tabular}{c|c|c c c} 
         
         \hline
        
        Domain & Metric &Shopping &Knowledge & Communication\\
        \hline
        Baseline& \multirow{3}{*}{RErr  (\%)} &0.0& 0.0	&0.0\\
        Oracle& &47.63&	40.28&	32.89\\
        PoolingAvg &&\textbf{25.55}&	\textbf{25.00}&	\textbf{11.92}\\
        \hline
        \end{tabular}}
        \vspace{-3ex}
\end{table}
Since another downstream task, intent classification, is similar to domain classification, we just show the best model in domain classification, PoolingAvg, on 
domain-specific
intent classification for three popular domains due to space limit. As Table \ref{tbl:intent} shows, the margins of using multiple hypotheses with PoolingAvg are significant as well.
\section{Conclusions and Future Work}
\label{sec:conclusion}
This paper improves the SLU system robustness to ASR errors by integrating $n$-best hypotheses in different ways, e.g. the aggregation of predictions from hypotheses or the concatenation of hypothesis text or embedding. 
We can achieve significant classification accuracy improvements over production-quality baselines on domain and intent classifications, 14\% to 25\% relative gains.  The improvement is more significant for a subset of testing data where ASR first best is different from transcription. We also observe that with more hypotheses utilized, the performance can be further improved. 
In the future, we aim to employ additional features (e.g. confidence scores for hypotheses or tokens) to integrate $n$-bests more efficiently, where we can train a function $f$ to obtain a weight for each hypothesis embedding before pooling. Another direction is using deep learning framework to embed the word lattice \cite{liu2014efficient} or confusion network \cite{hakkani2006beyond, tur2002improving}, which can provide a compact representation of multiple hypotheses and more information like times, in the SLU system.



\section{Acknowledgements}
We would like to thank Junghoo (John) Cho for proofreading.

\vfill\pagebreak

\bibliographystyle{IEEEbib}
\bibliography{main}

\end{document}